\documentclass[11pt]{article}

\usepackage[final]{acl}

\usepackage{times}
\usepackage{latexsym}
\usepackage{hyperref}
\usepackage{xurl}
\usepackage{booktabs}
\usepackage{makecell}
\usepackage{multirow}
\usepackage{tabularx}
\usepackage{tablefootnote}
\usepackage{url}
\usepackage{enumitem}
\usepackage{kotex}
\usepackage{amsmath}
\usepackage{tcolorbox}
\usepackage{array}

\newcommand{\sbullet}[1][0.75]{\mathbin{\vcenter{\hbox{\scalebox{#1}{$\bullet$}}}}}

\usepackage[T1]{fontenc}

\usepackage[utf8]{inputenc}

\usepackage{microtype}

\usepackage{inconsolata}

\usepackage{graphicx}

\title{HERIT: Enabling Global Access to Hanja-to-English Historical Translation by Mitigating Temporal Bias}

\author{
 \textbf{Dayeon Jeong\textsuperscript{1}},
 \textbf{Kyeongpil Kang\textsuperscript{2}}\thanks{Corresponding author}
\\
 \textsuperscript{1}Graduate School of Data Science, Kangwon National University \\
 \textsuperscript{2}Department of Computer Science and Engineering, Kangwon National University
\\
 \small{
   \{\href{mailto:glue0440@kangwon.ac.kr}{glue0440}, \href{mailto:rudvlf0413@kangwon.ac.kr}{rudvlf0413}\}@kangwon.ac.kr
 }
}

\begin{document}
\maketitle

\begin{abstract}
Historical archives are invaluable for multidisciplinary research but remain inaccessible globally due to severe language barriers. While manual translation is prohibitively expensive and time-consuming, simply applying existing machine translation models is inadequate due to unique linguistic nuances and the temporal bias inherent in scarce human-labeled corpora. To address these challenges, we propose HERIT, a systematic domain-adaptation pipeline that leverages retrieval-augmented generation (RAG) to generate high-quality pseudo-labeled data from abundant Hanja-Korean corpora. This approach substantially expands the training dataset, mitigating both data scarcity and temporal overfitting across different historical periods. Extensive automated and human evaluations demonstrate that HERIT consistently outperforms baselines. Finally, we translate previously untranslated portions of the archives using our model, thereby enabling global access to these resources for researchers.
\end{abstract}
\section{Introduction}
\label{sec:intro}
Historical records have been a vital source for discovering knowledge and deriving insights. In the humanities, these archives allow researchers to analyze political dynamics and cultural evolution. In the natural sciences, they serve as crucial sources for tracking natural phenomena, such as solar eclipses, supernovae, and seismic activities.

The Annals of the Joseon Dynasty (\textbf{AJD}) and the Journal of the Royal Secretariat (\textbf{JRS}) are regarded as preeminent historical archives, recognized not only in Korea but globally. They provide comprehensive records spanning over 500 years, comprising a vast array of subjects. Extending beyond state affairs and administration, they meticulously chronicle natural phenomena, including astronomical observations, seismic activity, famines, and epidemics. Accordingly, researchers worldwide utilize these archives across a broad spectrum of disciplines, from the social to the natural sciences.

Given their significance, the translation of the AJD from Hanja into Korean has been completed, while the translation of the JRS is currently underway. Furthermore, to promote the global accessibility of these historical archives, the National Institute of Korean History (NIKH) launched a project to translate the AJD into English. However, relying on manual translation by experts proficient in both Hanja and English presents significant challenges; the project is projected to span 22 years with a budget of approximately 20 million USD.\footnote{\url{https://sillok.history.go.kr/eslk/about/translationProjectInfo.do}} Furthermore, applying this manual approach to other archives such as the JRS, which is five times more voluminous than the AJD, would require prohibitive costs in both time and resources. Consequently, existing human-translated English corpora are not only scarce but also heavily skewed towards specific reigns. Improving translation on unseen historical periods requires a method that generalizes beyond this temporal bias.

Since these archives are written in Hanja, leveraging Chinese-English LLMs appears to be an intuitive solution for processing these corpora. However, such an approach encounters significant challenges. First, there is a fundamental script mismatch: Chinese models typically utilize Simplified Chinese, whereas Korean Hanja retains Traditional Chinese forms, resulting in character incompatibility. Second, substantial semantic and syntactic discrepancies exist. Even when characters are identical, semantic divergence is common, as homographs have often evolved distinct meanings in the respective languages. Furthermore, the texts contain Korea-specific Hanja usage that reflects native grammatical structures, and the word order frequently deviates from standard Chinese syntax. Most critically, Chinese-English models transliterate proper nouns, such as personal names and geographical locations, based on Chinese pronunciation rather than Korean pronunciation. This misalignment not only degrades translation quality but also introduces factual errors.

To address these challenges, we propose a systematic domain-adaptation pipeline for training a Hanja-to-English model specialized for historical archives, harnessing human-translated corpora, as shown in Fig.~\ref{fig:proposed_method}. To overcome the scarcity and temporal overfitting of available expert data, we introduce a retrieval-augmented in-context learning for data augmentation strategy that generates high-quality pseudo-labeled data. This effectively alleviates data sparsity while minimizing overfitting to specific eras. Subsequently, we present \textbf{HERIT}: a \textbf{H}anja-\textbf{E}nglish \textbf{R}AG-based \textbf{I}ntelligent \textbf{T}ranslation model for historical archives. Extensive evaluations, including quantitative metrics and human expert assessments, demonstrate that HERIT significantly outperforms existing baselines. In addition, we translate the remaining untranslated portions of the JRS and AJD using our model. Our model and the translated corpus are publicly available\footnote{\url{https://github.com/dandamdandam/HERIT}} to facilitate future research.

\section{Related Work}
\label{sec:related_work}
Historical archives have extensively contributed to diverse research domains. In the field of social sciences, they provide foundational data for analyzing international affairs~\cite{thies2002pragmatic}, demography~\cite{alter2019evolution}, economic conditions~\cite{clark2007long}, and linguistic transitions~\cite{wei2025diachronic}. In parallel, the natural sciences leverage historical records to investigate natural phenomena such as astronomical events~\cite{wang2021ancient,wei2024solar} and climate change~\cite{chen2020climate,zhang2021quantification}. Thus, historical documents serve as invaluable resources across diverse fields, offering a lens through which to understand past events and derive critical insights for both the present and the future.

Korean archives, such as AJD and JRS, are recognized as invaluable resources by the global academic community for their comprehensive scope, spanning multiple centuries and covering a wide array of subjects. Notably, the continuous and systematic management of these records makes them particularly advantageous for analyzing macroscopic changes. For instance, they have been extensively utilized in studies concerning astronomical observations and geological phenomena, as well as disasters such as epidemics, and even economic crises~\cite{stephenson2011historical,yang2012analysis,kim2017crustal,wang2021ancient,wei2024solar,rhee2014economic,hwang2022disastrous}.

With advances in machine learning, such as OCR and machine translation, research has been conducted to effectively translate historical archives into modern languages. \citet{afli2016integrating} proposed a method to digitize historical documents written in 17th-century French using OCR and translate them into Modern English by applying stochastic machine translation techniques. To translate Ancient Chinese into Modern Chinese, \citet{zhang2019automatic,liu2019ancient} developed passage-level alignment methods and deep neural network-based machine translation models.

Among Korean historical documents, the AJD has been fully translated into Korean by the NIKH, whereas the translation of the JRS remains ongoing due to its vast volume. Addressing this, \citet{kang2021restoring} developed a Hanja-to-Korean translation model leveraging the parallel corpus derived from both the AJD and JRS. Concurrently, a project to translate the original Hanja text of the AJD into English is underway, with the records from the era of King Sejong (approximately 2.5\% of the total) currently completed. Using this partial Hanja-English corpus alongside the Hanja-Korean data, \citet{son2022translating} proposed a Hanja-Korean-English multilingual model. However, due to the scarcity of human-translated Hanja-English data, the performance of models trained from scratch remains limited.

Recent LLMs, such as Gemini-2.5~\cite{comanici2025gemini}, Sonnet-4.5~\cite{anthropic2025claude-sonnet-4-5}, Kimi-K2~\cite{team2025kimi}, GPT-5.1~\cite{openai_gpt5_1}, and Qwen3~\cite{yang2025qwen3}, have demonstrated remarkable in-context learning capabilities. This advancement enables them to utilize information during the inference phase via few-shot learning, even without having encountered it during the pretraining or fine-tuning stages~\cite{brown2020language}. Specifically, RAG further improves performance by incorporating retrieved examples relevant to the input into the prompt~\cite{lewis2020retrieval}. Consequently, there is active research into data augmentation approaches for data-scarce domains that leverage these few-shot learning and RAG strategies~\cite{seo2024retrieval,song2024enhancing}. Moreover, methods such as Best-of-N (BoN)~\cite{liustatistical2024} and Fusion-of-N (FusioN)~\cite{khairi2025making} have been proposed to further enhance performance by generating multiple candidate responses and aggregating them to derive a final answer.

We address the scarcity of historical parallel corpora by leveraging RAG-based augmentation. Specifically, we construct a high-quality pseudo-labeled dataset using Hanja-Korean corpora via retrieval-augmented in-context learning and FusioN. Finally, we fine-tune a pre-trained LLM, leveraging its inherent English proficiency and partial Hanja understanding.
\section{Datasets}
\label{sec:datasets}

\begin{table}
    \centering
    \begin{tabular}{ccc}
        \toprule
        Hanja & Korean & English \\
        \midrule
        疫疾 & 역질 & Epidemic disease \\
        世宗 & 세종 & King Sejong \\
        江原道 & 강원도 & Kangwon Province \\
        \bottomrule
    \end{tabular}
    \caption{Examples of glossary information: Hanja source, Korean pronunciation, and English translation.}
    \label{tab:glossary_examples}
\end{table}

As noted in Section~\ref{sec:related_work}, the translation of the AJD is ongoing; currently, only the Annals of King Sejong have been fully translated. For our experiments, we obtained the AJD documents from NIKH\footnote{\url{https://www.history.go.kr/en/main/main.do}}.

Using these human-translated documents, we compiled a corpus yielding 350K Hanja-Korean pairs and 17.7K Hanja-Korean-English triples\footnote{Since all AJD documents were originally translated from Hanja to Korean, the Hanja-English data naturally forms (Hanja, Korean, English) triples.}. From the Hanja-English subset, we allocated 1,000 records each to the validation set $D^{\text{valid}}$ and the test set $D^{\text{test}}$, with the remaining 15.7K samples constituting the training set $D^{\text{train}}$. To mitigate period-specific evaluation bias, we constructed an additional test set for human evaluation, $D^{\text{test}}_{\text{NT}}$, by sampling 80 records per king (2,080 entries total) from the untranslated corpus. Crucially, $D^{\text{test}}_{\text{NT}}$ was strictly excluded from both the pseudo-labeling process and the training data.

To prevent data leakage, we implemented a rigorous decontamination pipeline. Given that Hanja is an ideographic script where individual characters carry dense semantic information, character-level exact and edit-based matching is particularly effective. Specifically, for each test instance, we retrieved the top-10 candidate matches from $D^{\text{train}}$ using BM25 and computed their character-level normalized Levenshtein similarity. We discarded any candidate exceeding a similarity threshold of $0.9$. We chose this threshold because historical archives frequently feature formalized administrative templates where changing a single entity or action drastically alters the translation semantics; lowering the threshold below $0.9$ would excessively discard valid, semantically distinct evaluation instances. This pipeline filtered out 83 samples from $D^{\text{test}}$, 3 from $D^{\text{test}}_{\text{NT}}$, and 1 from the 200-sample human evaluation subset, leaving the evaluation splits well-isolated from $D^{\text{train}}$. Detailed examples and similarity distribution for the deduplication process are provided in Appendix~\ref{sec:dedup}.

Using $D^{\text{train}}$ as an external knowledge base, we applied retrieval-augmented in-context learning for pseudo-labeling to the untranslated corpus (details in Section~\ref{subsec:building_a_pseudo_label_dataset}). After filtering out documents longer than 800 characters, we constructed an augmented dataset, $D^{\text{aug}}$, comprising 315K pseudo-labeled samples from all reigns except King Sejong's. Finally, both $D^{\text{train}}$ and $D^{\text{aug}}$ were utilized to train our model.

In addition, NIKH provides lexical resources, including semantic meanings and English pronunciations of proper nouns (e.g., personal and geographical names). Leveraging this data enables the LLM to comprehend Hanja documents more effectively and accurately translate proper nouns into English based on their correct pronunciation. Consequently, we compiled a glossary consisting of 25K entries from these lexicons and utilized it during the pseudo-labeling process.

\section{Proposed Methods}
\label{sec:proposed_methods}

\begin{figure*}[t]
    \centering
    \includegraphics[width=\linewidth]{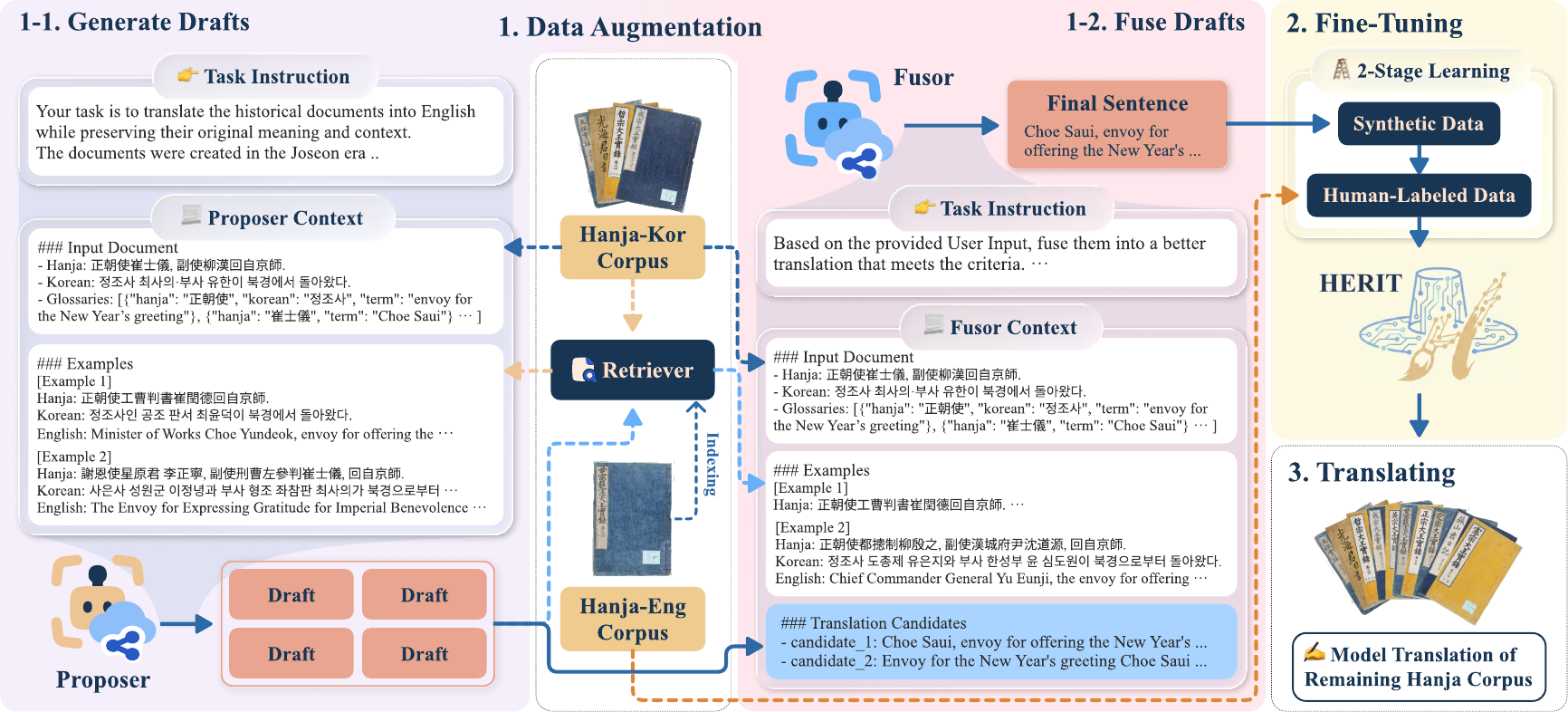}
    \caption{Overview of the proposed framework for translating historical documents into English. The framework comprises three main stages: 1) pseudo-labeling for data augmentation, 2) fine-tuning the LLM by integrating human-labeled and pseudo-labeled data, and 3) translation of the untranslated corpus using the trained model.}
    \label{fig:proposed_method}
\end{figure*}

This section presents our method for enhancing pseudo-label quality, followed by an introduction to our model fine-tuned on a blend of pseudo-labeled and expert-translated datasets.

\subsection{Selecting the Backbone LLM for Pseudo-Labeling}
\label{subsec:baseline_performance_of_llms}

Since the pseudo-labeling process can incur substantial API costs due to the large volume of target historical documents, evaluating candidate LLMs in terms of both performance and cost-effectiveness is essential for selecting the optimal backbone. To this end, we first assessed the translation performance of various LLMs on $D^{\text{test}}$ using base prompts. We employed standard metrics for machine translation, including BLEU~\cite{post-2018-call}, METEOR~\cite{banerjee2005meteor}, and chrF++~\cite{popovic2017chrf++}.

As shown in Table~\ref{tab:llm_base_performance}, Hunyuan-MT-7B, despite being a leading Chinese translation model~\cite{zheng2025hunyuan}, underperforms on Hanja translation. This result underscores the need for models specifically designed for Hanja. Results from the Qwen3 families demonstrate that translation quality generally improves as underlying model size increases. Regarding reasoning capabilities, while Gemini-2.5-Flash shows performance gains, Sonnet-4.5 experiences a slight decline. This aligns with prior surveys suggesting that reasoning does not significantly contribute to machine translation performance~\cite{zebaze2025llm}. A similar trend is observed with Gemini-2.5-Pro as shown in Table~\ref{tab:finetuning_performance}. Although Gemini-2.5-Pro achieves the highest accuracy, it incurs higher inference costs due to its size and reasoning overhead. In contrast, Gemini-2.5-Flash offers the most cost-effective trade-off, achieving competitive performance at a fraction of the cost. Therefore, we adopt Gemini-2.5-Flash for the pseudo-labeling process.

\setlength{\tabcolsep}{4pt}
\begin{table}[t]
    \centering
    \begin{tabular}{lccc}
        \toprule
        \makecell[c]{Model} & BLEU & chrF++ & Cost \$ \\
        \midrule
        Hunyuan-MT-7B & 8.318 & 35.450 & - \\
        Qwen3 8B & 10.631 & 38.738 & - \\
        Qwen3 32B & 12.062 & 40.461 & - \\
        GPT-5.1 & 13.574 & 43.545 & 1.92 \\
        Kimi-K2 (R) & 15.507 & 43.684 & 5.63 \\
        Gemini-2.5-Flash & 15.720 & 43.971 & 0.42 \\
        {\fontsize{10.8}{13}\selectfont Gemini-2.5-Flash (R)} & 17.183 & 45.878 & 3.06 \\
        Sonnet-4.5 & 18.131 & 48.254 & 5.39 \\
        Sonnet-4.5 (R) & 17.342 & 47.469 & 11.23 \\
        Gemini-2.5-Pro (R) & 19.640 & 48.926 & 18.17 \\
        \bottomrule
    \end{tabular}
    \caption{Translation performance on $D^{\text{test}}$ using basic prompts and the total API cost. (R) denotes the integration of reasoning capabilities.}
    \label{tab:llm_base_performance}
\end{table}

\subsection{Building a Pseudo-label Dataset for Hanja-English Translation}
\label{subsec:building_a_pseudo_label_dataset}

To improve translation performance, we construct a high-quality pseudo-labeled dataset by adopting retrieval-augmented in-context learning and FusioN~\cite{khairi2025making} approaches, following the process shown in Step 1 of Fig.~\ref{fig:proposed_method}. Specifically, in Step 1-1, the Proposer takes untranslated source documents to generate multiple translation drafts. In Step 1-2, the Fusor takes these drafts along with the original source document to generate the final translation. Since this architecture generates the final sentence by ensembling the Proposer's drafts in the Fusor, it not only mitigates mistranslation but also enhances translation quality by leveraging expanded information through a holistic examination of the drafts.

As depicted in Fig.~\ref{fig:proposed_method}, our goal is to improve the pseudo-label quality of the Proposer and Fusor by enriching the input prompt with auxiliary information, instead of relying solely on the Hanja document. To address the limited proficiency of LLMs in understanding Hanja, we augment the input with the corresponding Korean text to enhance contextual understanding. To ensure the accurate translation of proper nouns, we utilize a curated glossary by appending terms to the prompt when exact matches are found in the source text. We additionally utilize few-shot prompting to capture the nuances, including contextual and lexical information, of Hanja-to-English translation. Specifically, we adopt RAG that retrieves the top-$K$ similar documents $\mathcal{S}(h)$ from $D^{\text{train}}$ for a given input Hanja document $h$ via an embedding-based retriever. These retrieved entries are fed into both the Proposer and the Fusor as few-shot demonstrations in the form of (Hanja, Korean, English) triples.

We optimized the hyperparameters for pseudo-labeling based on the experiments described in Section~\ref{sec:pseudo-labeling}. To balance performance and inference cost, we selected Gemini-2.5-Flash as the backbone for both the Proposer and Fusor, and employed Gemini-Embedding~\cite{lee2025gemini} as the retriever. The retriever used the Korean component of the input document as the query for the Proposer, while utilizing both the input Hanja and the English translation drafts as the query for the Fusor. We employed 64 and 12 few-shot examples for the Proposer and the Fusor, respectively. During pseudo-labeling, the Fusor synthesized the final translation from five drafts generated by the Proposer with a temperature of 0.3.

Finally, to construct the pseudo-labeled dataset $D^{\text{aug}}$, we applied the proposed augmentation strategy to 317.4K Hanja-Korean pairs from corpora not yet translated into English (excluding data from the King Sejong era), incurring a cost of about \$3,000. This augmented dataset is utilized to mitigate data sparsity and prevent potential overfitting toward a specific historical period.

\subsection{Model Training and Translation of the Untranslated Corpus}
\label{subsec:model_training_and_translation}
We present our model, a Hanja-to-English historical document translation model fine-tuned on Qwen3 32B. Our model performs end-to-end translation from Hanja to English. Our training utilizes two datasets: a pseudo-labeled dataset $D^{\text{aug}}$ and a human-translated dataset $D^{\text{train}}$. We employ a two-stage training strategy: the first stage uses $D^{\text{aug}}$ to learn Hanja semantics and phonetic patterns of proper nouns, while the second stage leverages $D^{\text{train}}$ to align the model with human-level stylistic nuances. The entire training process was executed on eight H200 GPUs, taking 15.1 hours for the first stage and 1.2 hours for the second stage.

Finally, we deployed our model to translate all remaining untranslated AJD and JRS documents. Using eight NVIDIA H200 GPUs, the translation of these 2.09 million documents was completed in 94.6 hours, which is significantly faster and more cost effective than manual translation. To facilitate global accessibility to these historical records, both the dataset and the model are made publicly available.
\section{Experimental Results}
\label{sec:experimental_results}
In this section, we investigate optimal pseudo-labeling settings and demonstrate our model's superior performance against baselines including LLM-based and human expert evaluations.

\subsection{Ablation of the Pseudo-Labeling Pipeline}
\label{sec:pseudo-labeling}

\begin{table*}[t]
    \centering
    \begin{tabular}{cccccc}
        \toprule
        & & BLEU & METEOR & chrF++ & Cost \$ \\
        \midrule
        \multicolumn{2}{l}{Baseline} & 15.720 & 0.421 & 43.971 & 0.213 \\
        \multicolumn{2}{l}{+ Glossary} & 23.747 & 0.504 & 51.478 & 0.236 \\
        \multicolumn{2}{l}{+ Korean} & 25.457 & 0.520 & 52.637 & 0.255 \\
        \midrule
        \multicolumn{5}{l}{+ RAG ($K$: 12)} \\
        \multicolumn{2}{l}{\hspace{1em}$\sbullet$ Static shots} & 30.387 & 0.565 & 56.385 & 0.981 \\
        \multicolumn{2}{l}{\hspace{1em}$\sbullet$ Dynamic shots (Hanja)} & 36.223 & 0.604 & 60.354 & 1.115 \\
        \multicolumn{2}{l}{\hspace{1em}$\sbullet$ Dynamic shots (Kor)} & \textbf{36.430} & \textbf{0.606} & \textbf{60.545} & 1.254 \\
        \multicolumn{2}{l}{\hspace{1em}$\sbullet$ Dynamic shots (Hanja + Kor)} & 35.945 & 0.602 & 60.219 & 1.272 \\
        \midrule
        \multicolumn{2}{l}{RAG (Dynamic Shots, Kor, $K$: 64)} & 37.905 & 0.616 & 61.534 & 5.646 \\
        \midrule
        \multicolumn{5}{l}{+ FusioN ($K$: 12, $D$: 4)} \\
        \hspace{1em}Proposer (Kor) & Fusor (Kor) & 36.869 & 0.614 & 61.474 & 2.713 \\
        \hspace{1em}Proposer (Kor) & Fusor (Eng) & 36.901 & 0.613 & 61.420 & 2.475 \\
        \hspace{1em}Proposer (Kor) & Fusor (Hanja) & 36.985 & 0.614 & 61.538 & 2.537 \\
        \hspace{1em}Proposer (Kor) & Fusor (Hanja + Eng) & \textbf{37.017} & \textbf{0.614} & \textbf{61.504} & 2.642 \\
        \hspace{1em}Proposer (Kor) & Fusor (Hanja + Kor + Eng) & 36.926 & 0.614 & 61.499 & 2.806 \\
        \midrule
        \multicolumn{5}{l}{$K_\text{Proposer}$: 64, $K_\text{Fusor}$: 12, $D$: 5} \\
        \hspace{1em}Proposer (Kor) & Fusor (Hanja + Eng) & \textbf{38.462} & \textbf{0.625} & \textbf{62.501} & 7.059 \\
        \bottomrule
    \end{tabular}
    \caption{Ablation study of pseudo-labeling. The language in parentheses denotes the query language used for retrieval, while $K$ and $D$ represent the number of few-shot examples and translation drafts, respectively.}
    \label{tab:pseudo-labeling_ablation_test}
\end{table*}

To comprehensively validate the effectiveness of glossary integration, input augmentation, and few-shot prompting strategies, we evaluated translation performance on $D^{\text{test}}$ using Gemini-2.5-Flash. All experimental results with costs on Batch API are summarized in Table~\ref{tab:pseudo-labeling_ablation_test}.

First, we observe a significant improvement in translation performance when a glossary is incorporated. This suggests that the glossary information facilitates a better understanding of Hanja terms, aiding in their translation into appropriate English pronunciations. Similarly, providing pre-translated Korean inputs alongside the Hanja source text enhances the understanding of the input documents, indicating that Korean translations effectively support the pseudo-labeling process.

The latter part of the table presents the impact of RAG and FusioN, and the input language on the retriever. Including few-shot examples yields substantial performance gains. In particular, while static shots also improve performance, examples that are relevant to the input document lead to much larger gains, highlighting the effectiveness of RAG-based methods for enhancing performance. For dynamic shots, using the Korean translation of the input Hanja as the retriever query resulted in better performance than using the Hanja input alone. This is because the embedding model used for retrieval was trained on a larger corpus of Korean text compared to Hanja, resulting in superior comprehension of Korean.

Our method using FusioN outperforms standard single-stage translation pipelines. Experimental results with 12 few-shot examples and 4 drafts indicate that concatenating the Hanja input with the English draft sentences as query inputs for the retriever is effective when finding relevant documents for the Fusor. This suggests that in the Fusor, which refines drafts to generate the final translation output, both the context of the target Hanja sentence and that of the English candidate sentences are crucial. Additionally, we conducted experiments on the number of few-shots provided to the Proposer and Fusor, the results of which are shown in Fig.~\ref{fig:few_shot_for_fusion}. Finally, we set the number of drafts to 5 and few-shot examples for the Proposer and Fusor to 64 and 12, respectively.

As indicated in the bottom row of Table~\ref{tab:pseudo-labeling_ablation_test}, the final pseudo-labeling performance achieved a BLEU score of 38.462, outperforming other baselines, including the single translation approach that employs 64 dynamic few-shot examples (consistent with the proposer). Note that this performance does not reflect the direct translation of Hanja-only documents into English. Instead, it represents the pseudo-labeling performance facilitated by auxiliary information, such as Korean translations, glossary data, and retrieved references from $D^{\text{train}}$. Therefore, this rich auxiliary information enables the construction of a high-quality pseudo-labeled dataset, which subsequently improves model fine-tuning.

\subsection{Impact of Data Augmentation and Two-Stage Learning}

\begin{table}[t]
    \centering
    {\fontsize{10.4pt}{13pt}\selectfont
    \begin{tabular}{lccc}
        \toprule
        \makecell[c]{Model} & BLEU & METEOR & chrF++ \\
        \midrule
        H2KE & 24.62 & - & - \\
        Pivot-MT & 35.539 & 0.596 & 59.804 \\
        \midrule
        \multicolumn{4}{l}{With Glossary \& RAG} \\
        Gemini-2.5-Flash & 35.632 & 0.600 & 60.176 \\
        Gemini-2.5-Pro & 34.847 & 0.598 & 60.139 \\
        Qwen3 32B ($D^{\text{train}}$) & 38.117 & 0.622 & 62.338  \\
        \midrule
        \multicolumn{4}{l}{Without Glossary \& RAG} \\
        Qwen3 32B ($D^{\text{aug}}$) & 34.747 & 0.596 & 59.732 \\
        Qwen3 32B ($D^{\text{train}}$) & 36.930 & 0.613 & 61.583 \\        
        \midrule
        \multicolumn{4}{l}{+ Mixed training: $D^{\text{aug}} \cup D^{\text{train}}$} \\
        Qwen3 32B & 36.849 & 0.613 & 61.246 \\
        \midrule
        \multicolumn{4}{l}{+ Two-stage learning: $D^{\text{aug}} \rightarrow D^{\text{train}}$ (Ours)} \\
        Qwen3 8B & \textbf{37.849} & \textbf{0.623} & \textbf{62.187} \\
        Qwen3 32B & \textbf{38.432} & \textbf{0.626} & \textbf{62.525} \\
        \bottomrule
    \end{tabular}   
    }
    \caption{Translation performance on $D^{\text{test}}$. Our method is compared against baselines (H2KE~\cite{son2022translating} and Pivot-MT), zero-shot LLMs, and various ablation models. For H2KE, the BLEU score reported in the original paper is used.}
    \label{tab:finetuning_performance}
\end{table}

To verify the effectiveness of training with our augmented data, we evaluated translation performance using an expert-translated test dataset, $D^{\text{test}}$.

\begin{figure}[t]
    \centering
    \includegraphics[width=\columnwidth]{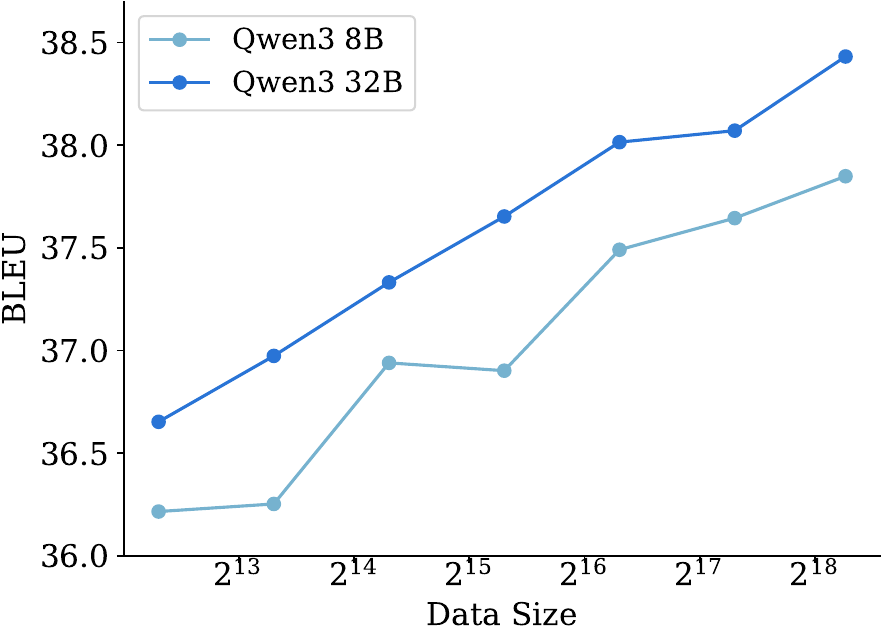}
    \caption{Translation performance with respect to pseudo-labeling data size.}
    \label{fig:scaling_pseudo_labeling}
\end{figure}

First, we evaluate the model performance with respect to the size of $D^{\text{aug}}$ used in the first training stage, as shown in Figure~\ref{fig:scaling_pseudo_labeling}. We observe that translation performance improves as the volume of pseudo-labeled data increases, demonstrating the effectiveness of our proposed approach.

Furthermore, we compared models fine-tuned exclusively on $D^{\text{train}}$ against those trained on both $D^{\text{train}}$ and $D^{\text{aug}}$. To isolate the effects of fine-tuning, we employed Gemini-2.5 Flash and Pro models utilizing glossary information and RAG as baselines. As shown in Table~\ref{tab:finetuning_performance}, given sufficient glossary context and relevant documents, Gemini-2.5-Pro demonstrates no significant performance advantage over Gemini-2.5-Flash. Although the model trained only on $D^{\text{train}}$ outperforms the one trained only on $D^{\text{aug}}$, the latter achieves performance comparable to Gemini-2.5-Pro augmented with a glossary and RAG. More importantly, incorporating our constructed $D^{\text{aug}}$ leads to better performance than training on $D^{\text{train}}$ alone, highlighting the effectiveness of our pseudo-labeling approach in low-resource historical document translation.

In addition, our two-stage training strategy outperforms the naive alternative of jointly training on a mixture of $D^{\text{aug}}$ and $D^{\text{train}}$. We attribute this to the severe size imbalance between the relatively small $D^{\text{train}}$ and the much larger $D^{\text{aug}}$, preventing the model from effectively leveraging both datasets when trained on them simultaneously. By contrast, the two-stage strategy allows the model to better utilize their complementary strengths. Furthermore, we fine-tuned an additional Qwen3-32B-based pivot translation baseline (Pivot-MT) under the same settings as our model. This baseline first translates Hanja into Korean and then translates the resulting Korean text into English. HERIT outperforms the Pivot-MT baseline, indicating that direct Hanja-to-English translation is more effective than pivot-based translation by avoiding error propagation across the two translation stages.

To examine whether pseudo-labeling can eliminate the need for inference-time retrieval, we fine-tuned Qwen3-32B on $D^{\text{train}}$ with a glossary and RAG applied during both training and inference. Although this configuration outperforms the $D^{\text{train}}$-only baseline, it still falls short of HERIT. More critically, retrieval augmentation incurs substantial computational overhead—expanding the context length to over 8K tokens (increasing the input prompt length by 13 to 14 times) and scaling training computational cost by 9 times. By contrast, HERIT effectively distills external knowledge directly into model parameters, achieving superior translation quality while remaining retrieval-free and highly efficient for large-scale historical text translation.

\subsection{LLM-Based and Human Evaluations}
\label{sec:human_evaluation}

\begin{table}[t]
    \centering
    \begin{tabular}{lcc}
        \toprule
        \makecell[c]{Model} & Qwen3 8B & Qwen3 32B \\
        \midrule
        \multicolumn{3}{l}{\textit{With Glossary \& RAG}} \\
        $D^{\text{train}}$ & 6.539 & 7.003 \\
        \midrule
        \multicolumn{3}{l}{\textit{Without Glossary \& RAG}} \\
        $D^{\text{train}}$ & 6.312 & 6.883 \\
        $D^{\text{aug}} \rightarrow D^{\text{train}}$ & \textbf{7.887} & \textbf{8.009} \\
        \bottomrule
    \end{tabular}
    \caption{LLM-based evaluation on the testset $D^{\text{test}}_{\text{NT}}$.}
    \label{tab:llm_based_evaluation}
\end{table}

Although $D^{\text{test}}$ consists of human-translated data, it is restricted to specific reign periods, which limits generalization across the entire chronological scope of the AJD corpus. To address this, we performed an LLM-based evaluation on the held-out dataset $D^{\text{test}}_{\text{NT}}$ by sampling 80 documents from each of the remaining reign periods excluded from pseudo-labeling. Adopting the method proposed by \citet{zheng2023judging}, models first generate translations for $D^{\text{test}}_{\text{NT}}$. Subsequently, the LLM Evaluator assesses the translation quality score on a scale of 1 to 10, taking the source Hanja text and the model-generated English translation as inputs. To enhance the reliability of model-based evaluation, we provided the LLM Evaluator with the corresponding human-translated Korean text and glossary information as reference materials. We chose Sonnet-4.6 as the evaluator to avoid potential bias arising from the training process.

As shown in Table~\ref{tab:llm_based_evaluation}, the LLM-based evaluation demonstrates that for both the 8B and 32B models, the two-stage learning approach combining pseudo-labeled and human-translated data outperforms models trained exclusively on human-translated data. Notably, given that $D^{\text{test}}_{\text{NT}}$ covers the entire AJD period disjoint from the training dataset, these results confirm the effectiveness of pseudo-labeling in enhancing generalization.

Furthermore, while incorporating the glossary and RAG into both training and inference slightly improves performance over the baseline trained solely on $D^{\text{train}}$, the two-stage model trained on $D^{\text{aug}}$ substantially outperforms the model fine-tuned with the glossary and RAG. This highlights that pseudo-labeling provides significant additional performance gains, demonstrating that our approach effectively generalizes across the entire AJD period rather than temporal overfitting to the specific timeframe of $D^{\text{train}}$.

\begin{figure}[t]
    \centering
    \includegraphics[width=\columnwidth]{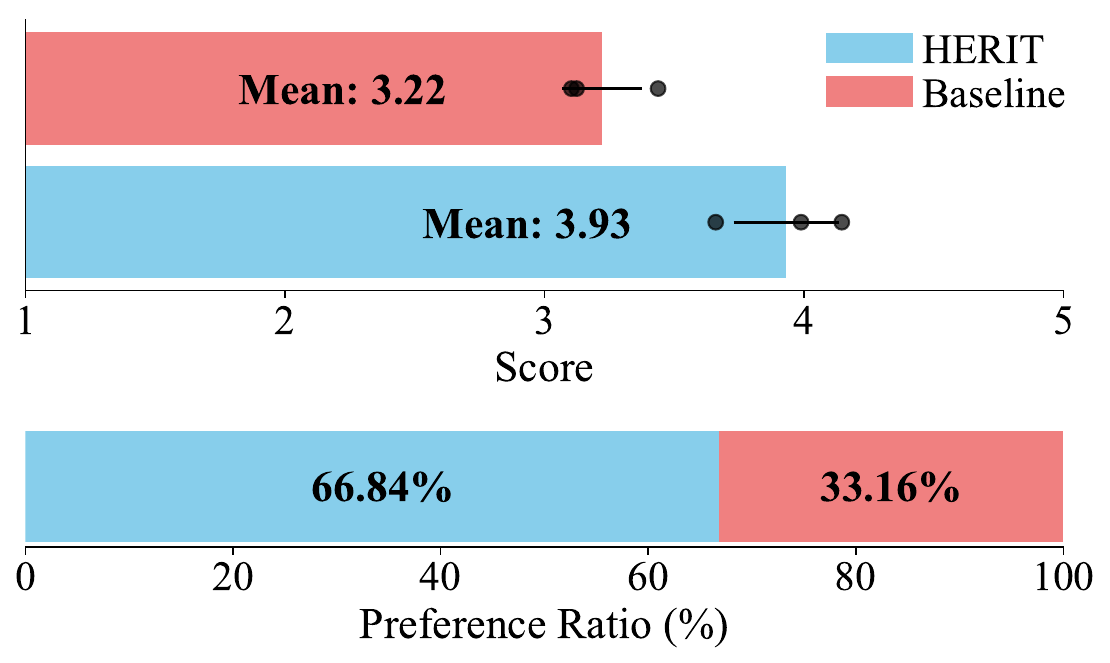}
    \caption{Human expert evaluation scores and preference ratios for our model versus the baseline.}
    \label{fig:expert_eval}
\end{figure}

Additionally, using 200 examples sampled uniformly across different kings from $D^{\text{test}}_{\text{NT}}$, we recruited domain experts in Sinographic literature and Korean history who are proficient in English to evaluate our model against the baseline (Qwen3 32B trained solely on $D^{\text{train}}$). We asked the experts to rate each translation on a scale of 1-5. Further, to differentiate between models in cases where scores were tied (excluding identical translations), we performed a comparative evaluation where experts were instructed to select the better response.

As shown in Fig.~\ref{fig:expert_eval}, our model achieved a score of 3.93, outperforming the baseline score of 3.22. In addition, HERIT achieved a selection rate of 66.84\% in the pairwise preference evaluation, which is more than twice that of the baseline. These results demonstrate the superior translation quality of HERIT in human expert evaluation.

\section{Analysis}
This section evaluates the model's robustness against temporal overfitting and provides a qualitative analysis of translation examples.

\subsection{Temporal Overfitting}
\label{sec:temporal_overfitting}

\begin{figure*}[t]
    \centering
    \includegraphics[width=\linewidth]{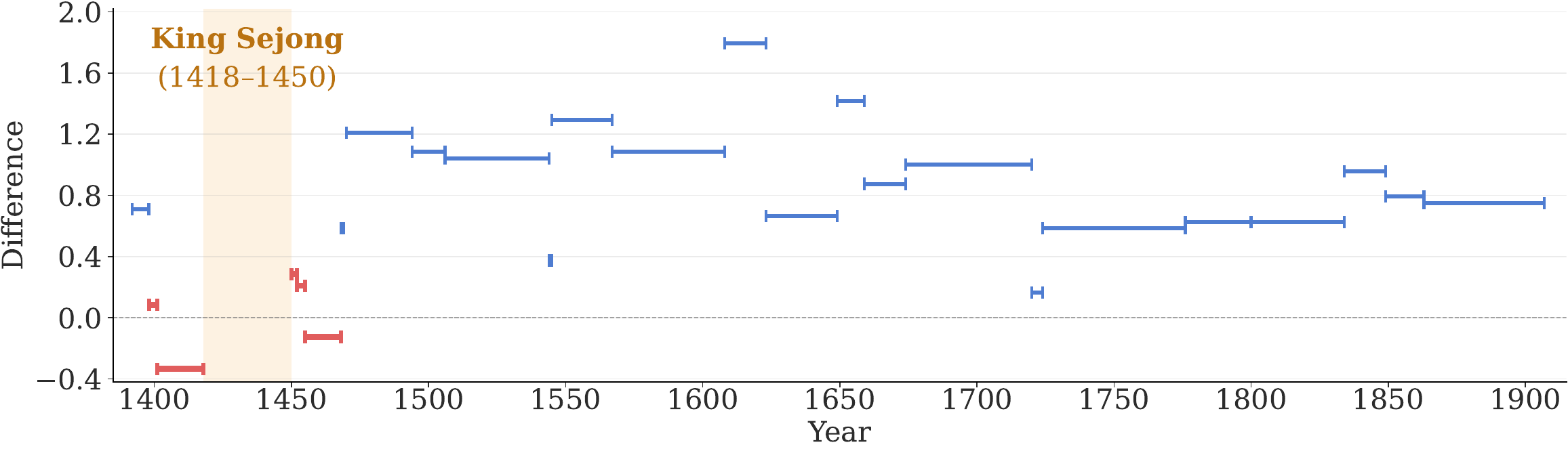}
    \caption{Differences in human expert scores between HERIT (ours) and Qwen3 32B trained solely on human-labeled data, ordered chronologically by each king's reign. Positive values indicate that HERIT outperforms the baseline, demonstrating its robustness across different historical periods.}
    \label{fig:diff_between_pseudo_human}
\end{figure*}

To investigate whether our method effectively mitigates temporal overfitting, we analyzed the human evaluation results. Since the evaluation samples were stratified by king, as detailed in Section~\ref{sec:human_evaluation}, we computed the average score for each reign across both HERIT and the baseline, Qwen3 32B trained exclusively on human-labeled data. We then determined the performance gap between the two models for each reign and visualized these gaps chronologically.

As noted in Section~\ref{sec:related_work}, the human-labeled corpus is heavily concentrated on documents from the reign of King Sejong. Consequently, Qwen3 trained on this temporally biased dataset performs comparably to HERIT during this specific period, as shown in Fig.~\ref{fig:diff_between_pseudo_human}. In contrast, HERIT consistently achieves superior performance on documents from other reigns. This indicates that training on data concentrated in a single historical period leads to temporal overfitting, while the proposed pseudo-labeling strategy helps mitigate this bias and enhances overall translation robustness.

\subsection{Qualitative Analysis of Translations}
\begin{table*}[t]
\setlength\doublerulesep{0.15cm}
\centering
{\fontsize{11pt}{12.8pt}\selectfont
  \begin{tabularx}{\textwidth}{l|X}
  \Xhline{2\arrayrulewidth}
  Input Hanja & 兩司啓趙應奎事, 不允. \\
  \hline
  \multirow{2}{*}{\makecell[l]{Hunyuan-MT}} & a petition was submitted by the two departments regarding \textcolor{red}{Zhao Yingkui}’s case, but it was denied. \\
  \hline
  \multirow{2}{*}{\makecell[l]{Qwen3 32B}} & The two investigative bodies reported on the matter of \textcolor{blue}{Jo Eunggyu}, but the king did not approve their request. \\
  \hline
  HERIT & The two offices reported on the matter of \textcolor{blue}{Jo Eunggyu}, but the king did not approve. \\
  \hline\hline
  Input Hanja & 宜給宣飯及房子, 炊飯, 汲水人等 \\
  \hline
  Hunyuan-MT & It is appropriate to provide food, \textcolor{red}{housing, cooking facilities, and water for the people.} \\
  \hline
  Qwen3 32B & They should be provided with food, \textcolor{red}{a house}, \textcolor{blue}{a female palace servant}, and a water carrier. \\
  \hline
  HERIT & We should provide them with meals, \textcolor{blue}{female palace servants}, cooks, and water carriers. \\  
  \hline\hline
  Input Hanja & 宥魚思漢, 京外從便 \\
  \hline
  Hunyuan-MT & \textcolor{red}{Yu Yu thinks of Han;} outside the capital, \textcolor{red}{do as you please.} \\
  \hline
  \multirow{2}{*}{\makecell[l]{Qwen3 32B}} & The king pardoned \textcolor{blue}{Eo Sahan} and allowed him to \textcolor{blue}{reside in a place of his choice} \textcolor{red}{in the capital} or in the provinces. \\
  \hline
  \multirow{2}{*}{\makecell[l]{HERIT}} & The king pardoned \textcolor{blue}{Eo Sahan} and allowed him to \textcolor{blue}{reside in a place of his choice outside the capital.} \\
  \Xhline{2\arrayrulewidth}
\end{tabularx}
}
\caption{Qualitative examples of translations by different models using samples in $D^{\text{test}}$. Red denotes incorrect translations, while blue indicates correct ones.}
\label{tab:translation_examples}
\end{table*}

We also conducted a qualitative comparison of HERIT against two baselines using samples from the test set: Qwen3 32B fine-tuned exclusively on human-translated data and Hunyuan-MT-7B, a leading Chinese translation model.

As shown in the first example in Table~\ref{tab:translation_examples}, both Qwen3 32B and HERIT correctly transliterated the proper nouns according to Korean pronunciation rules, whereas Hunyuan-MT-7B incorrectly relied on Chinese phonetics. In the second case, the term {\fontsize{10pt}{12.5pt}\selectfont 房子} carries distinct semantics depending on the context: it denotes a ``house'' in standard Chinese but refers to a ``female palace servant'' in the context of the Joseon Dynasty. While Hunyuan-MT-7B defaulted to the standard meaning and Qwen3 32B conflated the interpretations, HERIT identified the historical context and translated the term correctly. In the last example, Hunyuan-MT-7B failed to recognize the proper name {\fontsize{10pt}{12.5pt}\selectfont 魚思漢}, and neither baseline recognized the penal term {\fontsize{10pt}{12.5pt}\selectfont 京外從便} (a form of punishment). In contrast, HERIT accurately translated these domain-specific terms.

These results demonstrate HERIT’s superior ability to deliver precise translations by capturing both Sino-Korean lexical nuances and the unique contexts of historical documents.

\section{Conclusion}
In this study, we presented HERIT, a systematic domain-adaptation pipeline designed to address the scarcity of parallel corpora in historical translation. By leveraging abundant intermediate resources through a RAG- and Fusion-based strategy, we developed a robust method for generating high-quality pseudo-labels without relying on costly human annotation. Extensive experiments, including both human and LLM-based evaluations, demonstrated that HERIT not only outperforms strong baselines but also effectively alleviates temporal overfitting, a persistent challenge in analyzing historical archives. Furthermore, our results highlight the potential of LLM-based data augmentation for other low-resource classical languages where direct translation pairs are limited. By enabling the translation of previously inaccessible documents, our work facilitates broader access to historical archives, empowering researchers worldwide to engage more deeply with historical texts.

\section*{Limitations}
Due to budgetary constraints, we primarily employed Gemini-2.5-Flash for the Proposer and Fusor components, focusing exclusively on AJD documents for pseudo-labeling. We anticipate that employing a more diverse set of LLMs as Proposers to enhance translation diversity, as well as incorporating JRS documents as targets, would yield higher-quality pseudo-labels. Furthermore, hardware limitations necessitated restricting the maximum sequence length to 2,048 tokens, with a limit of 800 Hanja characters. Consequently, approximately 6.7\% of the training data was filtered out. For inference, documents exceeding this limit were processed using a chunking strategy. We expect that extending the model’s maximum input length will further improve translation performance.

\section*{Acknowledgments}
This work was supported by the National Research Foundation of Korea(NRF) grant funded by the Korean government(MSIT) in 2026 (RS-2025-00555320, Contribution Rate: 70\%; RS-2023-00242528, Regional Demand Oriented Data Science Human Resource Development for Regional Industry Innovation, Contribution Rate: 20\%), the ANCHOR program through the Gangwon ANCHOR Center, funded by the Ministry of Education(MOE) and Gangwon State(G.S.), Republic of Korea(2026-ANCHOR-10-002, Contribution Rate: 10\%), and the Google Cloud Research Credits program with the award GCP19980904.

\bibliography{reference}

\clearpage

\appendix

\section{Deduplication}
\label{sec:dedup}

\begin{table*}[t]
    \centering
    \begin{tabularx}{\textwidth}{l l X X}
    \toprule
    Case & Language & Sample in $D^{\text{test}}$ & Searched Sample in $D^{\text{train}}$ \\
    \midrule
    \multirow{2}{*}{\shortstack[l]{Example 1 \\ (Sim: 0.900)}} 
    & Hanja & 宗貞盛遣人來獻\textcolor{blue}{土宜}. & 宗貞盛遣人來獻\textcolor{blue}{土物}. \\
    \cmidrule(lr){2-4}
    & English & Sō Sadamori sent a messenger to present local products. & Sō Sadamori sent a man to present his local products to the king. \\
    \midrule
    \multirow{2}{*}{\shortstack[l]{Example 2 \\ (Sim: 0.857)}} 
    & Hanja & 上王詣健元陵, 行\textcolor{blue}{端午}別祭. & 上王詣健元陵, 行\textcolor{blue}{中秋}別祭. \\
    \cmidrule(lr){2-4}
    & English & The abdicated king went to the royal tomb of King Taejo and held a special ancestral rite of \textcolor{blue}{the fifth day of the fifth month}. & The abdicated king visited the royal tomb of King Taejo to conduct the special sacrificial rite for \textcolor{blue}{the Mid-Autumn}. \\
    \bottomrule
    \end{tabularx}
\caption{Qualitative examples of test–train sample pairs with high Levenshtein similarity.}
\label{tab:deduplication_examples}
\end{table*}


\begin{figure}[t]
    \centering
    \includegraphics[width=\columnwidth]{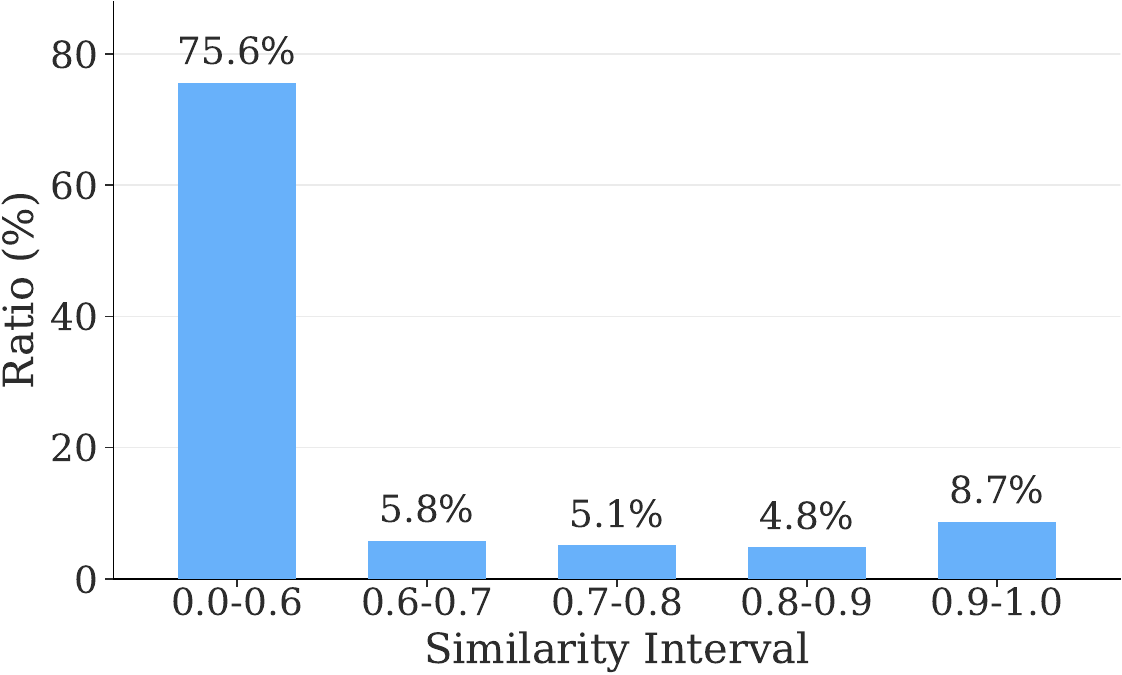}
    \caption{Distribution of test set samples across similarity intervals against the training data.}
    \label{fig:deduplication_similarity_ratio}
\end{figure}

To justify our similarity threshold, Table~\ref{tab:deduplication_examples} presents qualitative comparisons around the threshold. In Example 1, the sentence essentially repeats a standardized tribute record with near-identical phrasing and semantics, justifying its removal. Conversely, Example 2 demonstrates why a lower threshold would over-filter; replacing the ritual name (端午 vs. 中秋) radically shifts the historical context, date, and English translation despite high lexical overlap. Manual inspection across intervals confirmed that 0.9 strictly eliminates near-duplicate administrative phrasing while preserving semantically distinct formulaic records.

Figure~\ref{fig:deduplication_similarity_ratio} shows the distribution of maximum Levenshtein similarity between $D^{\text{test}}$ and $D^{\text{train}}$. The majority of evaluation samples ($75.6\%$) exhibit low overlap ($\le 0.6$). Crucially, no significant clustering is observed near the 0.90 boundary, ensuring that the benchmark remains well-isolated from training data and genuinely assesses generalization rather than template recall.

\section{Training Details}
\label{sec:training_details}
We trained the models for one epoch in each stage. To mitigate overfitting and prevent a bias toward the later stage, we employed stage-specific learning rates. For the 32B model, the learning rates were set to $1\times10^{-5}$ for the first stage and $5\times10^{-6}$ for the second; for the 8B model, we applied $2\times10^{-5}$ and $1\times10^{-5}$, respectively. The batch size was set to 64 for the first stage and 16 for the second stage. All models were trained with the PagedAdamW 8-bit optimizer~\cite{dettmers2022optimizers} and FSDP~\cite{zhao2023pytorch} on eight NVIDIA H200 GPUs.

\section{Experimental Results}
\subsection{Number of Few-shot Examples}

\begin{figure}[t]
    \centering
    \includegraphics[width=\columnwidth]{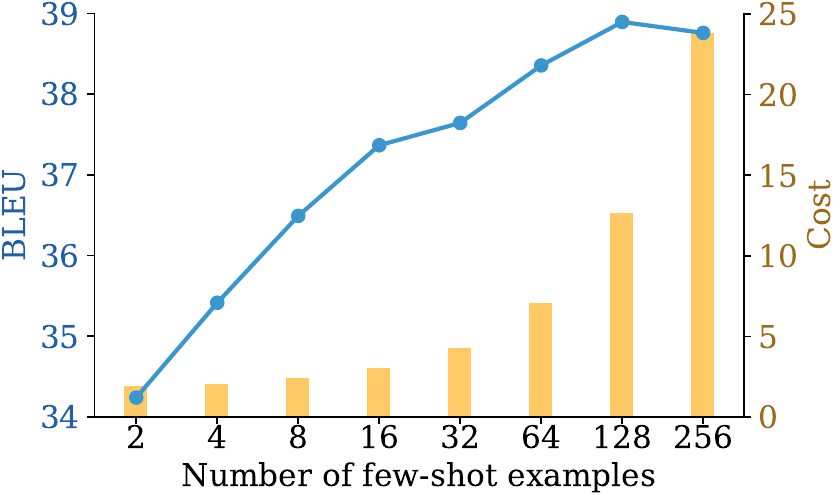}
    \includegraphics[width=\columnwidth]{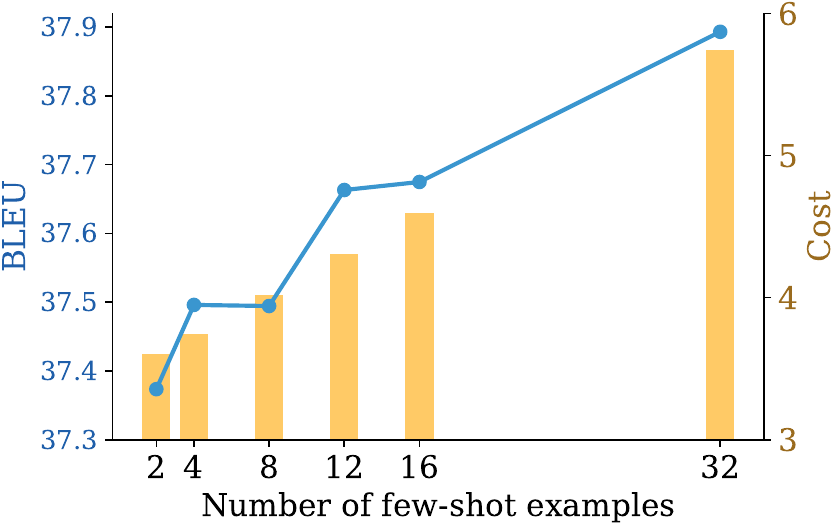}
    \caption{Performance and cost of the Proposer and Fusor across different numbers of few-shot examples.}
    \label{fig:few_shot_for_fusion}
\end{figure}

Fig.~\ref{fig:few_shot_for_fusion} illustrates the impact of the number of few-shot examples on the performance of both the Proposer and Fusor. Notably, the Proposer exhibits consistent performance gains up to 128 examples. However, to balance performance against the increased LLM inference costs incurred by larger shot sizes, we adopted a 64-shot configuration for the Proposer during pseudo-labeling. Additionally, we identified that a temperature of 0.3 and a draft count of 5 provided the best performance, and thus utilized these parameters for pseudo-labeling. In contrast, although the Fusor also improves with more examples, its marginal gains are less pronounced than those of the Proposer. Therefore, considering the performance-cost trade-off, we set the few-shot size for the Fusor to 12.

\subsection{Impact of FusioN}

\setlength{\tabcolsep}{5.5pt}
\begin{table}[t]
    \centering
    \begin{tabular}{lccc}
        \toprule
         & BLEU & METEOR & chrF++ \\
        \midrule
        No aggregation & 37.905 & 0.616 & 61.534 \\
        \midrule
        BoN & 38.337 & 0.622 & 62.117 \\
        FusioN & \textbf{38.462} & \textbf{0.625} & \textbf{62.501} \\
        \bottomrule
    \end{tabular}
    \caption{Comparison of pseudo-labeling performance among FusioN, BoN, and Baseline.}
    \label{tab:comparison_fusion_bon}
\end{table}

We also evaluated the pseudo-labeling performance of multi-candidate aggregation methods, specifically FusioN and BoN. As shown in Table~\ref{tab:comparison_fusion_bon}, BoN outperforms the baseline, which generates translations in a single stage without candidate aggregation. However, FusioN demonstrates superior performance compared to BoN, suggesting that synthesizing a final output by aggregating all candidates is more effective than simply selecting the single best candidate.

\setlength{\tabcolsep}{3.8pt}
    \begin{table}[t]
    \centering
    \begin{tabular}{lcccc}
        \toprule
        \makecell[c]{Method} & BLEU & METEOR & chrF++ & Cost (\$) \\
        \midrule
        Multiple & 36.250 & 0.613 & 61.507 & 234.304 \\
        Single & \textbf{38.462} & \textbf{0.625} & \textbf{62.501} & \textbf{58.517} \\
        \bottomrule
    \end{tabular}
    \caption{Performance comparison between diverse-model and single-model FusioN approaches.}
    \label{tab:fusion_diverse_models}
\end{table}

In addition, we investigated the impact of proposer diversity on cost-effectiveness. The single-proposer setting relied solely on Gemini-2.5-Flash, whereas the diverse-proposer setting combined Qwen3 235B-A22B, DeepSeek-V3.1, GPT-5.1, Gemini-2.5-Flash, and Sonnet 4.5. Table~\ref{tab:fusion_diverse_models} reports the performance and cost of the single- and diverse-proposer settings, taking prompt caching into account. The results show that the single-model configuration achieves both better performance and higher cost efficiency. This suggests that while increasing model diversity yields a wider variety of candidates, it does not necessarily translate into better performance, making the single-model configuration a more cost-effective choice.

To qualitatively verify that FusioN effectively synthesizes proposer-generated translations during the aggregation stage, we analyze the representative examples shown in Table~\ref{tab:fusion_case_studies}. In the first example, while candidate (2) misinterprets the syntactic structure, yielding a flawed translation, FusioN successfully adopts the correct phrasing from candidate (1), preserving semantic accuracy. In the second example, candidate (2) suffers from incomplete generation. FusioN effectively filters out this noise and consolidates valid information from the viable candidates to produce a fluent and comprehensive translation. In the third example, although multiple candidates generate the generic term ``white falcon,'' FusioN selects the historically accurate term ``white goshawk'' provided by a minority candidate, demonstrating its capability for fine-grained lexical selection. Overall, these cases illustrate that FusioN effectively aggregates candidate translations to ensure both factual correctness and grammatical fluency.

\begin{table*}[t]
    \centering
    \begin{tabularx}{\textwidth}{c|X}
        \Xhline{2\arrayrulewidth}
        Input Hanja & 御勤政殿受朝. \\
        \hline
        Ground Truth & The King proceeded to Geunjeong Hall and received officials in the morning court assembly. \\
        \hline
        \multirow{3}{*}{\makecell[c]{Translation \\ by Candidates}} & (1) The king proceeded to Geunjeong Hall and \textcolor{blue}{received officials in the morning court assembly}. \\
        & (2) The king proceeded to Geunjeong Hall to \textcolor{red}{receive the morning court assembly}. \\
        & ... \\
        \hline
        \makecell[c]{Translation \\ by Fusor} & The king proceeded to Geunjeong Hall and received officials in the morning court assembly. \\
        
        \hline \hline
        Input Hanja & 傳旨兵曹: 自今司僕寺常養馬, 冬節則三百匹, 夏節則二百匹. \\
        \hline
        \multirow{3}{*}{Ground Truth} & The King had a royal directive transmitted to the Ministry of War: ``Hereafter, the number of horses to be reared continually by the Office of Royal Stables shall be 300 head in winter and 200 head in summer.'' \\
        \hline
        \multirow{5}{*}{\makecell[c]{Translation \\ by Candidates}} & 
        (1) The King transmitted a royal directive to the Ministry of War as \textcolor{blue}{follows: ``From this day forward, the Office of the Royal Stables shall always raise horses, with 300 horses in winter and 200 horses in summer''} \\
        & (2) The King transmitted a royal directive to the Ministry of War as \textcolor{red}{follows:} \\
        & ... \\
        \hline
        \multirow{3}{*}{\makecell[c]{Translation \\ by Fusor}} & The King transmitted a royal directive to the Ministry of War, stating,``From this day forward, the Office of the Royal Stables shall always raise horses, with 300 horses in winter and 200 horses in summer'' \\
        \hline \hline
        Input Hanja & 咸吉道監司, 進白鷹一連. \\
        \hline
        Ground Truth & Governor of Hamgil Province presented the king with one white goshawk. \\
        \hline
        \multirow{4}{*}{\makecell[c]{Translation \\ by Candidates}} & (1) The governor of Hamgil Province presented the king with one \textcolor{blue}{white goshawk}. \\
        & (2) The governor of Hamgil Province presented the king with one \textcolor{red}{white falcon}. \\
        & (3) The governor of Hamgil Province presented a \textcolor{red}{white falcon} to the king. \\
        & ... \\
        \hline
        \makecell[c]{Translation \\ by Fusor} & The Governor of Hamgil Province presented the king with one white goshawk. \\
        \Xhline{2\arrayrulewidth}
    \end{tabularx}
    \caption{Qualitative case studies showing how FusioN synthesizes candidate translations.}
\label{tab:fusion_case_studies}
\end{table*}

\subsection{Translation Analysis for Proper Nouns}

\begin{table}[t]
    \centering
    \begin{tabular}{lc}
        \toprule
        \makecell[c]{Model} & Accuracy \\
        \midrule
        Qwen3 32B ($D^{\text{train}}$) & 0.818 \\
        \midrule
        Qwen3 32B ($D^{\text{aug}} \cup D^{\text{train}}$) & 0.875 \\
        \midrule
        Qwen3 32B ($D^{\text{aug}} \rightarrow D^{\text{train}}$) & \textbf{0.878} \\
        \bottomrule
    \end{tabular}
    \caption{LLM-based evaluation of proper noun accuracy across the baseline, pseudo-labeling, and training strategies.}
    \label{tab:compare_proper_noun_acc}
\end{table}

To assess whether our proposed pseudo-labeling and two-stage training strategy effectively capture historical and factual context, we conducted an LLM-based evaluation. In particular, we examined whether the model correctly renders proper nouns using their actual Korean pronunciations rather than Chinese readings. Specifically, using Claude Sonnet-4.6, we evaluated the translation accuracy of proper nouns—including historical figures, geographical names, and official titles—in $D^{\text{test}}$. As shown in Table~\ref{tab:compare_proper_noun_acc}, the proposed pseudo-labeling method substantially improves proper-noun translation accuracy over the baseline model trained only on $D^{\text{train}}$. Furthermore, the two-stage learning strategy provides additional gains beyond those achieved by pseudo-labeling alone.

\section{Human Evaluation}

\begin{figure}[t]
    \centering
    \includegraphics[width=\columnwidth]{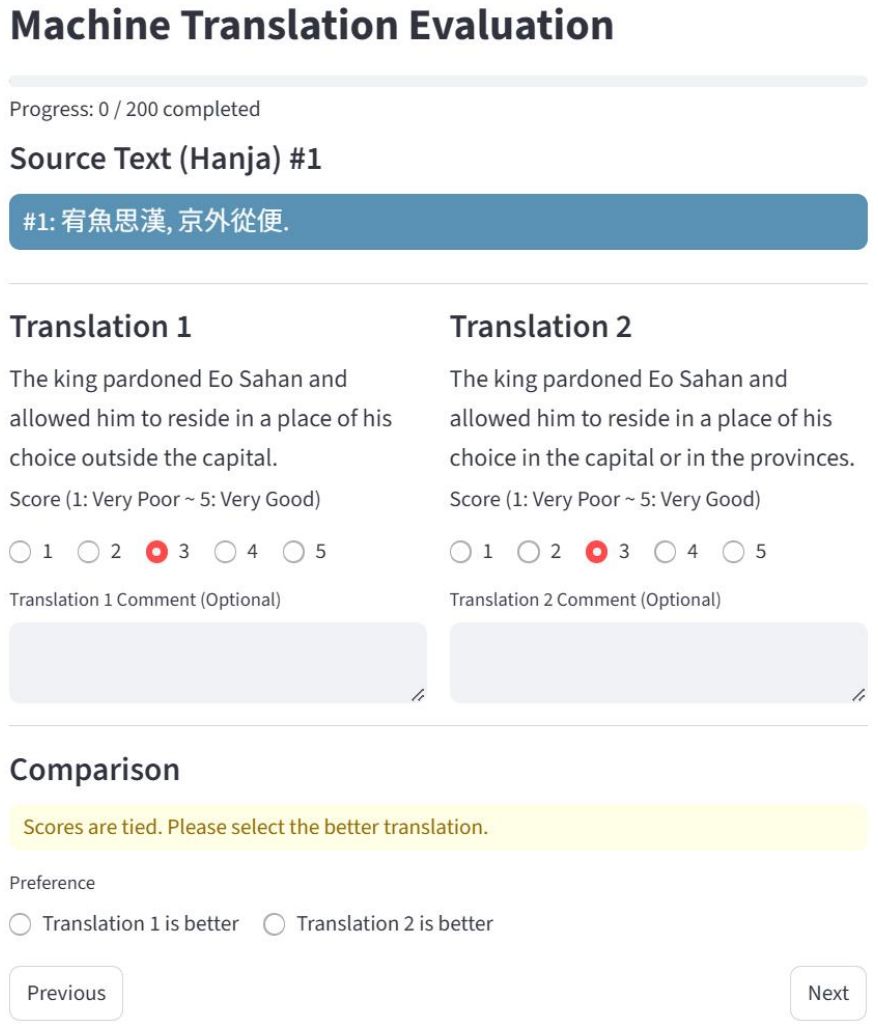}
    \caption{Web-based evaluation tool for human experts.}
    \label{fig:expert_eval_tool_interface}
\end{figure}

To compare our model against the baseline, we recruited three professional experts in Sinographic literature and Korean history who are proficient in English to evaluate 200 items. The evaluation was conducted in a strict double-blind manner, ensuring that the evaluators remained completely unaware of the identity of the generating models. Prior to the task, the experts were provided with comprehensive instructions regarding the interface and the evaluation rubric.

In addition, we developed a custom web-based interface to facilitate this evaluation, as shown in Fig.~\ref{fig:expert_eval_tool_interface}. The interface displays the source Hanja text and the associated glossary at the top, while the anonymized outputs from both models are presented in the center. Evaluators rated each translation on a 1-5 Likert scale. To ensure a rigorous comparative analysis, a forced-choice mechanism was implemented; whenever a tie occurred (i.e., identical scores were given to both models), evaluators were required to select the superior translation based on three core criteria:
\begin{itemize}[nosep]
    \item \textbf{Semantic coherence}: The degree of preservation of the original meaning without omission or distortion.
    \item \textbf{Terminological consistency}: The correct and uniform translation of historical terms and proper nouns.
    \item \textbf{Naturalness}: The fluency and readability of the translated English text.
\end{itemize}

Along with the quantitative scores, evaluators provided qualitative comments justifying their preferences or pointing out specific errors in the counterpart translations, including semantic distortion, information omission, terminology misuse, and unnatural phrasing.

\section{Prompt Format}
Fig.~\ref{fig:translation_prompt} shows the base prompt template used for direct Hanja-to-English translation in the baseline models discussed in Sections~\ref{subsec:baseline_performance_of_llms} and~\ref{sec:experimental_results}. The system prompt defines the instructions for the translation task and provides a description of Hanja within the ``Task'' section. It also outlines ``Guidelines for translation'' and briefly explains the user prompt structure and expected output format. In the user prompt, the source Hanja text and relevant glossary information are provided in the ``Input Document'' section. For RAG-based baselines, relevant ``Examples'' are included as (Hanja, English) pairs.

Similarly, Figs.~\ref{fig:proposer_prompt} and \ref{fig:fusor_prompt} present the prompt structures used in the pseudo-labeling process described in Section~\ref{subsec:building_a_pseudo_label_dataset}. Unlike the base prompt, the Proposer and Fusor prompts incorporate Korean translations provided by human experts, which serve as guidance for translating historical documents. The Fusor prompt largely follows the structure of the Proposer prompt, but differs in several key respects. Following the implementation of \citet{khairi2025making}, it includes a ``Task'' section describing the draft-fusion method and a ``Fusion Steps'' section outlining the procedure. In the user prompt, the task instructions and guidelines from the Proposer stage are included under ``Translation Instruction,'' while the drafts generated by the Proposer are listed under ``Translation Candidates.'' The Fusor leverages these translation candidates to generate the final translation.

\newenvironment{promptitemize}
  {\begin{itemize}[itemsep=0pt, topsep=0pt, parsep=0pt, leftmargin=*]}
  {\end{itemize}}

\newenvironment{promptbox}
{
  \begin{tcolorbox}[
    width=\linewidth,
    colback=white,
    colframe=black,
    boxrule=1pt,
    arc=2mm,
    left=4pt,
    right=4pt,
    top=5pt,
    bottom=5pt,
    fonttitle=\bfseries
  ]
  \small
}
{
  \end{tcolorbox}
}

\definecolor{mypink1}{rgb}{0.0941, 0.2118, 0.3294}

\newenvironment{promptsubbox}[1]
{
  \begin{tcolorbox}[
    width=\linewidth,
    colback=white,
    colframe=mypink1,
    boxrule=0.6pt,
    arc=2mm,
    left=4pt,
    right=4pt,
    top=5pt,
    bottom=5pt,
    title=\textbf{#1},
    fonttitle=\bfseries\small
  ]
  \small
}
{
  \end{tcolorbox}
}

\begin{figure*}[t]
\centering
\begin{promptsubbox}{System Prompt}

\textbf{\#\#\# Task}

Your task is to translate the historical documents into English while preserving their original meaning and context.  
The documents were created in the Joseon era in Korea (from the late 14th century to the early 20th century) and were written in Hanja, a traditional writing system used at the time.

\vspace{0.5em}
\textbf{\#\#\# Guidelines}

Focus on accurately conveying the meaning, nuances, and context of the original texts.

\vspace{0.5em}
\textbf{\#\#\# Input}

The document contains the original Hanja text, and may optionally include glossaries or examples of expert translations.

\vspace{0.5em}
\textbf{\#\#\# Output JSON Format}

\texttt{\{``english\_translation'': str\}}

\end{promptsubbox}
\begin{promptsubbox}{User Prompt}
\textbf{\#\#\# Examples}

[Example 1]\\
Hanja: 上王幸豐壤, 觀離宮之役而還. 離宮乃豐壤縣古基, 去都城四十里.\\
English: The abdicated king visited Pungyang, observed the construction of a detached palace, and returned. The detached ...

\vspace{0.3em}

[Example 2]\\
Hanja: 上王幸離宮南郊, 觀放鷹.\\
English: The abdicated king went to the southern outskirts of the temporary palace and viewed falconry.

...

\vspace{0.5em}
\textbf{\#\#\# Input Document}

\begin{promptitemize}
  \item hanja: 上王幸毋岳, 觀離宮造成之役.
  \item glossaries: [\{``hanja'':``上王'', ``korean'': ``상왕'', ``term'': ``Abdicated King''\}, \{``hanja'': ``毋岳'', ``term'': ``Muak''\}, ...]
\end{promptitemize}

\end{promptsubbox}
\vspace{-0.3cm}
\caption{An example of the base prompt.}
\label{fig:translation_prompt}
\end{figure*}

\begin{figure*}[t]
\centering
\begin{promptsubbox}{System Prompt}

\textbf{\#\#\# Task}

Your task is to translate the historical documents into English while preserving their original meaning and context.  
The documents were created in the Joseon era in Korea (from the late 14th century to the early 20th century) and were written in Hanja, a traditional writing system used at the time.

\vspace{0.5em}
\textbf{\#\#\# Guidelines}

Focus on accurately conveying the meaning, nuances, and context of the original texts.

\vspace{0.5em}
\textbf{\#\#\# Input}

The document contains the original Hanja text, its Korean translation, and may optionally include glossaries or examples of expert translations.

\vspace{0.5em}
\textbf{\#\#\# Output JSON Format}

\texttt{\{``english\_translation'': str\}}

\end{promptsubbox}
\begin{promptsubbox}{User Prompt}
\textbf{\#\#\# Examples}

[Example 1]\\
Hanja: 正朝使工曹判書崔閏德回自京師.\\
Korean: 정조사인 공조 판서 최윤덕이 북경에서 돌아왔다.\\
English: Minister of Works Choe Yundeok, envoy for offering the New Year's greetings to the Emperor of Ming China ...

\vspace{0.3em}

[Example 2]\\
Hanja: 謝恩使星原君 李正寧, 副使刑曹左參判崔士儀, 回自京師.\\
Korean: 사은사 성원군 이정녕과 부사 형조 좌참판 최사의가 북경으로부터 돌아왔다.\\
English: The Envoy for Expressing Gratitude for Imperial Benevolence, Lord of Seongwon Yi Jeongnyeong, and ...

...

\vspace{0.5em}
\textbf{\#\#\# Input Document}

\begin{promptitemize}
  \item Hanja: 正朝使崔士儀, 副使柳漢回自京師.
  \item Korean: 정조사 최사의·부사 유한이 북경에서 돌아왔다.  
  \item Glossaries: [\{``hanja'': ``正朝使'', ``korean'': ``정조사'', ``term'': ``envoy for the New Year's greeting''\}, \\ \hspace*{4.8em} \{``hanja'': ``崔士儀'', ``term'': ``Choe Saui''\} ... ]
\end{promptitemize}

\end{promptsubbox}
\vspace{-0.3cm}
\caption{An example of the prompt format used in the Proposer.}
\label{fig:proposer_prompt}
\end{figure*}

\begin{figure*}[!t]
\centering
\begin{promptsubbox}{System Prompt}

\textbf{\#\#\# Task}

Based on the provided User Input, fuse them into a better translation that meets the criteria.  
The source texts are Joseon-era documents originally written in Hanja between the late 14th and early 20th centuries. You must create a fused translation that faithfully preserves the original meaning, nuances, and historical context.

\vspace{0.5em}
\textbf{\#\#\# User Input}

\begin{promptitemize}
  \item Examples: (reference material) Certified example translations.
  \item Input Document
  \begin{promptitemize}
    \item Hanja: The original Hanja text.
    \item Korean: Its Korean translation.
    \item Glossaries: (reference material) Glossaries to clarify specific words and characters.
  \end{promptitemize}
  \item Translation Instruction: Additional instructions for translation.
  \item Translation Candidates: Multiple English translation candidates.
\end{promptitemize}

\vspace{0.5em}
\textbf{\#\#\# Fusion Steps}

\begin{enumerate}[itemsep=0pt, topsep=1pt, parsep=0pt, leftmargin=*]
  \item Read all source materials and translation candidates.
  \item Compare strengths and weaknesses.
  \item Generate a fused final translation.
\end{enumerate}

\vspace{0.5em}
\textbf{\#\#\# Output JSON Format}

\texttt{\{``final\_translation'': str\}}

\end{promptsubbox}
\begin{promptsubbox}{User Prompt}
\textbf{\#\#\# Examples}

[Example 1]  \\
Hanja: 正朝使工曹判書崔閏德回自京師.  \\
Korean: 정조사인 공조 판서 최윤덕이 북경에서 돌아왔다.  \\
English: Minister of Works Choe Yundeok, envoy for offering the New Year's greetings to the Emperor of Ming China ...

\vspace{0.3em}

[Example 2]  \\
Hanja: 正朝使都摠制柳殷之, 副使漢城府尹沈道源, 回自京師.  \\
Korean: 정조사 도총제 유은지와 부사 한성부 윤 심도원이 북경으로부터 돌아왔다.  \\
English: Chief Commander General Yu Eunji, the envoy for offering the New Year's greetings to the Emperor of Ming ...

...

\vspace{0.5em}
\textbf{\#\#\# Input Document}

\begin{promptitemize}
  \item Hanja: 正朝使崔士儀, 副使柳漢回自京師.
  \item Korean: 정조사 최사의·부사 유한이 북경에서 돌아왔다.
  \item Glossaries: [\{``hanja'': ``正朝使'', ``korean'': ``정조사'', ``term'': ``envoy for the New Year's greeting''\}, \\ \hspace*{4.8em} \{``hanja'': ``崔士儀'', ``term'': ``Choe Saui''\} ... ]
\end{promptitemize}

\vspace{0.5em}
\textbf{\#\#\# Translation Instruction}

Task: Translate the historical documents into English while preserving their original meaning and context.  

Guidelines: Focus on accurately conveying the meaning, nuances, and context of the original texts.

\vspace{0.5em}
\textbf{\#\#\# Translation Candidates}

\begin{promptitemize}
  \item Candidate\_1: Envoy for the New Year's greeting Choe Saui and Vice-Envoy Yu Han returned from Beijing.
  \item Candidate\_2: Choe Saui, envoy for the New Year's greeting, and Yu Han, vice-envoy, returned from Beijing.
  \item Candidate\_3: Choe Saui, envoy for offering the New Year's greetings to the Emperor of Ming China, and Yu Han ...
\end{promptitemize}
\hspace*{0.5em}...
\vspace{0.2em}

\end{promptsubbox}
\vspace{-0.3cm}
\caption{An example of the prompt format used in the Fusor.}
\label{fig:fusor_prompt}
\end{figure*}

\section{Use of AI Assistants}
During the preparation of this manuscript, the authors used AI assistants, such as Gemini and Claude, for proofreading, grammar correction, and improving readability. All core ideas, implementations, experimental evaluations, and analyses were developed by the authors.

\end{document}